\documentclass[letterpaper]{article} 
\usepackage{aaai24}  
\usepackage{times}  
\usepackage{helvet}  
\usepackage{courier}  
\usepackage[hyphens]{url}  
\usepackage{graphicx} 
\usepackage{natbib}  
\usepackage{times}  
\usepackage{helvet}  
\usepackage{courier}  
\usepackage[hyphens]{url}  
\usepackage{graphicx} 

\usepackage{times}
\usepackage{soul}
\usepackage{url}
\usepackage{multirow}
\usepackage{multicol}
\usepackage{amsfonts}
\usepackage[utf8]{inputenc}

\usepackage{amsmath}
\usepackage{amsthm}
\usepackage{booktabs}
\usepackage{algorithm}
\usepackage{algorithmic}
\usepackage{subfig}

\usepackage{color}
\usepackage{graphicx}
\usepackage{booktabs}

\usepackage{caption} 
\usepackage{algorithm}
\usepackage{algorithmic}

\usepackage{newfloat}
\usepackage{listings}
\DeclareCaptionStyle{ruled}{labelfont=normalfont,labelsep=colon,strut=off} 
\floatstyle{ruled}
\newfloat{listing}{tb}{lst}{}
\floatname{listing}{Listing}
\title{Exploiting  Multi-Label Correlation in Label Distribution Learning}
\author{
    Zhiqiang Kou\textsuperscript{\rm 1}\textsuperscript{\rm 2}\equalcontrib, Jing Wang\textsuperscript{\rm 1}\textsuperscript{\rm 2}\equalcontrib, 
    Yuheng Jia\textsuperscript{\rm 1}\textsuperscript{\rm 2}\thanks{Corresponding author.} and
    Xin Geng\textsuperscript{\rm 1}\thanks{Corresponding author.}\\
    }
\affiliations{
    \textsuperscript{\rm 1} School of Computer Science and Engineering, Southeast University, Nanjing 210096, China\\

\textsuperscript{\rm 2} MOE Key Laboratory of Computer Network and Information Integration, China\\

\{zhiqiang \_Kou, wangjing91, yhjia, xgeng \}@seu.edu.com,
    
}

\usepackage{bibentry}

\begin{document}

\maketitle

\begin{abstract}
	Label Distribution Learning (LDL) is a novel machine learning paradigm that assigns label distribution to each instance. Many LDL methods proposed to leverage label correlation in the learning process to solve the exponential-sized output space; among these, many exploited the low-rank structure of label distribution to capture label correlation. However, recent studies disclosed that label distribution matrices are typically full-rank, posing challenges to those works exploiting low-rank label correlation. Note that multi-label is generally low-rank; low-rank label correlation is widely adopted in multi-label learning (MLL) literature. Inspired by that, we introduce an auxiliary MLL process in LDL and capture low-rank label correlation on that MLL rather than LDL. In such a way, low-rank label correlation is appropriately exploited in our LDL methods. We conduct comprehensive experiments and demonstrate that our methods are superior to existing LDL methods. Besides, the ablation studies justify the advantages of exploiting low-rank label correlation in the auxiliary MLL. 
\end{abstract}

\section{Introduction}

\begin{figure}[!h]
	\centering
	\includegraphics[width=0.45\textwidth]{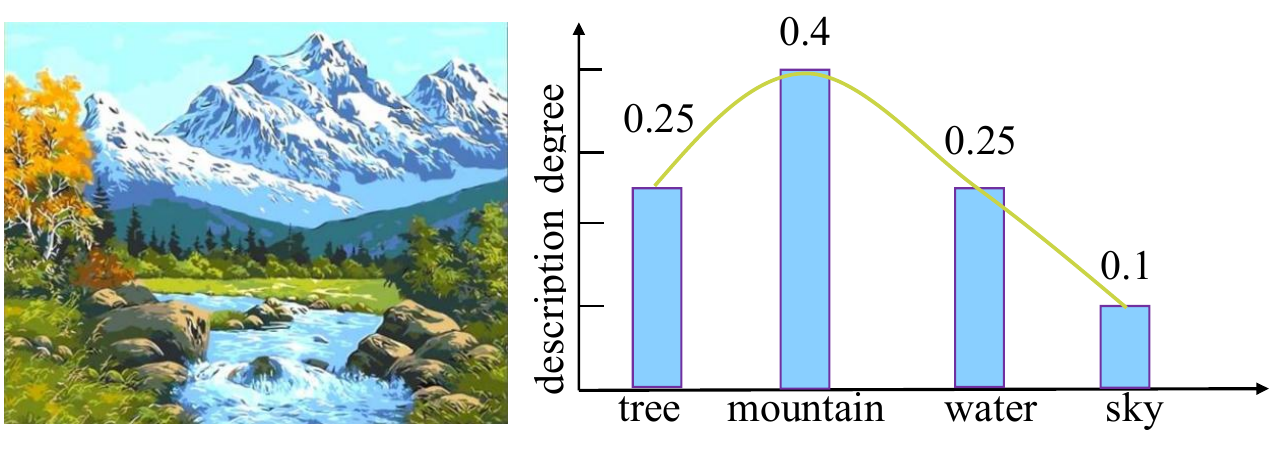} 
	\caption{An image from a natural-scene dataset \cite{geng2014multilabel} with a label
		distribution. }
	\label{tu1}
\end{figure}

Label distribution learning (LDL) \cite{geng2016label} is a novel learning paradigm that provides fine-grained label information for each instance.  Unlike traditional learning paradigms, LDL introduces the label description degree \cite{geng2016label} that is a real-value and quantify the relevance of one labels to a specific example. The label description degrees of all labels form a label distribution, which provides a comprehensive representation of labeling information. Fig.\ref{tu1} showcases an image from a natural-scene \cite{geng2014multilabel} dataset.
The average ratings are rescaled to form a label distribution $\{0.25, 0.4, 0.25, 0.1\}$, effectively capturing the varying degrees of importance assigned to labels.  LDL utilizes label description degrees to effectively solve label ambiguity \cite{gao2017deep}. 

LDL has an overwhelming output space, which exponentially grows with the number of potential labels \cite{wang2021label_2}. To tackle this issue, label correlation has emerged as a promising solution. 
Many LDL works have proposed to leverage label correlation in the learning processes. To name a few, \cite{xu2017incomplete} captured the low-rank structure of label distribution matrix to incorporate label correlation. 
\cite{jia2019label} leveraged the low-rank assumption to capture label correlations shared by different groups of samples. They effectively captured label correlations in a local context. 
Additionally, \cite{ren2019label} utilized a low-rank matrix to capture the global label correlation and further updated it based on different clusters to also explore local label correlations. Label correlation helps significantly improve the performance of these LDL methods. 
Notice that the above mentioned LDL methods assumed that label distribution has low-rank structure and relied it to exploit label correlation. However, \cite{wang2021label_2} has demonstrated that label distribution matrices are usually full-rank, which challenges the suitability of these approaches with low-rank assumption. So, it is possible to exploit the low-rank label correlation in LDL more efficiently?

Our work is mainly inspired by two observations. The first one is that low-rank has been extensively applied to multi-label learning (MLL) to capture label correlation \cite{xu2016robust}\cite{jing2015semi}\cite{liu2021emerging}\cite{wu2020joint}\cite{xu2016robust}\cite{yu2018feature}. The second one is that label distribution has rich supervision information and implicitly contains multi-label information. To see that, for the example in Fig.~\ref{tu1} we can observe that the given label distribution contains implicit multi-labels of \{\textit{tree}, \textit{mountain}, \textit{water}\}. Our basic idea is to introduce an auxiliary MLL process in LDL and exploit the low-rank label correlation on the MLL part, for example, by assuming the MLL matrix is low-rank. That is, the low-rank assumption is added on the MLL part instead of the LDL part.  

Following the strategy, we propose two novel LDL methods TLRLDL and TKLRLDL to exploit low-rank label correlation. First, we propose two methods to generate multi-label from label distribution. The first one utilizes a threshold to separate label distribution into multi-label, and the second one selects the top-$k$ labels having the largest label description degrees as the positive labels. Next, we learn label distribution and the generated multi-label simultaneously, and capture the low-rank label correlation in the MLL process. We conduct expensive experiments to justify that the proposed methods outperform existing LDL approaches. Besides, the ablation studies validate the advantages of exploiting the low-rank label correlation in the MLL process. To sum up, our major contributions are as follows:
\begin{itemize}
	\item As far as we know, this is the first work to introduce an auxiliary MLL process in LDL and exploit MLL label correlation for LDL. 
	
	\item We exploit label correlation in LDL by capturing the low-rank structure in an auxiliary MLL process, which is more reasonable than directly exploiting the low-rank label correlation of LDL.  
	
	\item We conduct extensive experiments to validate the advantages of our methods over existing LDL algorithms and the superiority of exploiting the low-rank label correlation in the auxiliary MLL process.
\end{itemize}

\section{Related Work}
\label{related-work}

\subsection{Lable Distribution Learning}
As a novel learning paradigm, LDL introduces label distribution to accurately capture the degree of labeling for each instance. This unique characteristic has sparked significant research interest in the field of LDL. In this section, we provide a brief overview of the existing studies in LDL. 

The existing LDL \cite{le2023uncertainty}\cite{tan2023label} algorithms can be broadly classified into three categories. The first category involves transforming the LDL problem into a single-label learning problem by assigning weights to the training samples. Representative algorithms in this category include PT-SVM and PT-Bayes, which utilize SVM and Bayes classifiers to solve the transformed weighted single-label learning problem.  The second category focuses on adapting traditional machine learning algorithms to handle the LDL problem. For instance, the K-nearest neighbors (KNN) classifier identifies the top k neighbors of an instance and predicts the label distribution by averaging the labels of these neighbors. Back-Propagation (BP) neural networks directly optimize the descriptive degree of the final prediction through the BP algorithm. The third category comprises specialized algorithms such as IIS-LDL and BFGS-LDL. These algorithms formulate LDL as a regression problem and employ improved iterative scaling and quasi-Newton methods, respectively, to solve the regression problem efficiently. However, these LDL algorithms do not take label correlation into account.

\subsection{Label Correlation in LDL}
In recent years, researchers have recognized the challenge of the vast output space of LDL \cite{gao2017deep}\cite{wang2019classification}\cite{shen2017label}\cite{wang2021label}\cite{ren2017sense} and have developed various approaches to address this issue. These methods can be categorized into three main types: 1) global label correlation, 2) local label correlation, and 3) both global and local label correlations. In the first category, \cite{zhou2015emotion} introduced a weighted Jeffrey's divergence \cite{cha2007comprehensive} to capture label correlation by assigning weights based on the Pearson correlation coefficient. \cite{xu2017incomplete} incorporated the low-rank structure of label distribution by applying trace-norm regularization. In the second category, \cite{jia2019label} utilized a local low-rank structure to implicitly capture the local label correlations.  In the third category,  \cite{ren2019label} introduced LDL-LCLR, a method that leverages both global and local label correlations. It utilizes a low-rank matrix to capture global label correlation and updates the matrix based on different clusters to explore local label correlation. 

However, many of these works rely on the assumption of low-rank to exploit label correlation. As reported by \cite{wang2021label_2}, label distribution matrix is generally full rank, which poses a challenge to those exploiting low-rank label correlation of LDL. Instead of directly exploiting low-rank label correlation of LDL, this study introduces an auxiliary MLL and exploits the low-rank label correlation on the MLL process, which can efficiently solve the above mentioned problem.

\section{The Proposed Method }

\textbf{Notations:}  Let $\mathbf{X}=\left[\mathbf{x}_{1}, \mathbf{x}_{2}, \ldots, \mathbf{x}_{n}\right] \in \mathbb{R}^{n \times d}$ denote the feature matrix and $\mathbf{Y}=\left\{y_{1}, y_{2}, \ldots, y_{m}\right\}$ be the label space,  where $n$, $m$, and $d$  denote the numbers of instances, labels, and the dimension of features, respectively. 
The training set of the LDL is represented as $\mathbf{T} =\left\{\left(\mathbf{x}_{1}, \mathbf{d}_{1}\right), \quad\left(\mathbf{x}_{2}, \mathbf{d}_{2}\right), \ldots,\left(\mathbf{x}_{n}, \mathbf{d}_{n}\right)\right\}$, where $\mathbf{d}_{i}=\left[d_{\mathbf{x}_{i}}^{y_{1}}, d_{\mathbf{x}_{i}}^{y_{2}}, \ldots, d_{\mathbf{x}_{i}}^{y_{m}}\right]$ is the label distribution of the $i$th sample $\mathbf{x}_{i}$.  $d_{\mathbf{x}_{i}}^{y}$ is the label description degree of $y$ to $\mathbf{x}_{i}$, which satisfies  $d_{\mathbf{x}_i}^{y} \in[0,1]$ and $\sum_{y} d_{\mathbf{x}}^{y}=1$. The label distribution matrix is denoted as $\mathbf{D}=\left[\mathbf{d}_{1}, \mathbf{d}_{2}, \ldots, \mathbf{d}_{n}\right] \in \mathbb{R}^{m \times n}$.  LDL aims to learn a mapping function from $\mathbf{T}$ and predict the label distribution for unseen instances. 

\subsection{Transforming Label Distribution into Multi-Label}
This subsection will introduce two methods for transforming label distribution into multi-label. The first one is threshold-based degradation, and the second one is top-$k$ degradation. Next, we provide details of these two methods.

\subsubsection{Threshold-Based Multi-Label Generation}
To convert label distribution into multi-label, we simulate the labeling process that users typically follow when assigning labels to images or adding keywords to texts. Overall, users continue adding the most relevant labels until they perceive that the labeling is sufficiently comprehensive\cite{xu2019label}. Based on that, we can degrade multi-label from label distribution through this iterative labeling procedure. The process is outlined as follows:
\begin{itemize}
	\item For each instance $\mathbf{x}$, find the label $y_j$ with the highest description degree $d_{\mathbf{x}}^{y_j}$ and add it to relevant labels (i.e., $l_{\mathbf{x}}^{y_j}=1$).
	\item Calculate the sum of the description degrees of all the currently relevant labels $H=\sum_{y_j \in \mathcal{Y}^{+}} d_{\boldsymbol{x}}^{y_j}$, where $\mathcal{Y}^{+}$ is the set of the currently relevant labels.
	\item  If $H$ is less than a predefined threshold $T$, continue finding the label with the highest description degree from the labels not included in $\mathcal{Y}$, and add it to $\mathcal{Y}$. Repeat this process until $H>T$.
\end{itemize}
Following this process, we can generate multi-label from label distribution that mimics the way users label data. 

\subsubsection{Top-$k$ Based Multi-Label Generation}
Specifically, for any instance $\mathbf{x}_i$, we first sort the label description degrees in descending order. Then, we select the top-$k$ labels with the highest label description degrees as relevant labels, and assign the remaining labels as irrelevant labels. That is, the top-$k$ labels with the highest label description degrees are considered relevant for each instance, while the rest are deemed irrelevant.

\subsection{Auxiliary MLL and Label Correlation}

First, we employ the least square method to learn label distribution and minimize the $L_2$-norm loss between the ground-truth label distribution and prediction, which can be formalized as the following:
\begin{equation}
	\min _\mathbf{W} \frac{1}{2}\left\|\mathbf{W} \mathbf{X}^{\top}-\mathbf{D}\right\|_F^2+\lambda\|\mathbf{W}\|_F^2,
	\label{EQ1}
\end{equation}
where $\mathbf{W}\in \mathbb{R}^{m \times d}$ is the parameter matrix, $\|\cdot\|_{F}$ represents the Frobenius norm, and $\lambda$ is a regularization parameter. Next, we establish the mapping relationship between label distribution and multi-label generated in the previous section. This linear mapping is formulated as:
\begin{equation}
	\min _\mathbf{O}\|\mathbf{D O-L}\|_F^2+ \lambda \|\mathbf{O}\|_F^2,
	\label{eq2}
\end{equation}
where $\mathbf{O}\in \mathbb{R}^{n \times n}$ is the transformation parameter matrix. 

Next, we exploit the low-rank label correlation. However, given the full-rank nature of the label distribution matrix, assuming a low-rank structure does not suit LDL. To address that, we encourage the low-rank structure on the MLL process, which has been widely accepted in MLL literature. That is, the predicted MLL matrix is assumed to be low-rank, which further casts Eq. (\ref{eq2}) as:
\begin{equation}
	\min _\mathbf{O}\|\mathbf{D O-L}\|_F^2+\alpha \text{Rank}(\mathbf{W}\mathbf{X}^{\top}\mathbf{O})+\eta\|\mathbf{O}\|_F^2,
	\label{eq3}
\end{equation}
where $\text{Rank}(\mathbf{A})$ represents the rank of $\mathbf{A}$, and $\alpha$ is a balance parameters.  By jointly optimizing Problem (1) and Problem (3), we obtain the final formulation as follows:
\begin{equation}
\small
	\begin{gathered}
		\min _{\mathbf{W,O}} \frac{1}{2}\left\|\mathbf{W} \mathbf{X}^{\top}-\mathbf{D}\right\|_F^2+\frac{1}{2}\left\|\mathbf{W} \mathbf{X}^{\top} \mathbf{O}-\mathbf{L}\right\|_F^2+\\
		\alpha \text{Rank}\left(\mathbf{W}\mathbf{X}^{\top} \mathbf{O}\right) 
		+\lambda\left(\|\mathbf{W}\|_F^2+\|\mathbf{O}\|_F^2\right),
	\end{gathered}
	\label{EQ3}
\end{equation}

$\text{Rank}(\mathbf{W})$ is difficult to solve due to the discrete nature of the rank function. Fortunately, as suggested by \cite{candes2011robust}, the nuclear-norm \cite{fazel2002matrix} is a good surrogate for the rank function. Replacing the rank function with the nuclear-norm, we obtain the next optimization problem: 
\begin{equation}
	\begin{gathered}
		\min _{\mathbf{W,O}} \frac{1}{2}\left\|\mathbf{W} \mathbf{X}^{\top}-\mathbf{D}\right\|_F^2+\frac{1}{2}\left\|\mathbf{W} \mathbf{X}^{\top} \mathbf{O}-\mathbf{L}\right\|_F^2+\\
		\alpha\left\|\mathbf{W}\mathbf{X}^{\top} \mathbf{O}\right\|_{*}  
		+\lambda\left(\|\mathbf{W}\|_F^2+\|\mathbf{O}\|_F^2\right). 
	\end{gathered}
	\label{EQ4}
\end{equation}
The method learning multi-label by threshold is called TLRLDL and the other one learning multi-label from top-$k$ is called TKLRLDL. 

\subsection{Optimization}
We use  ADMM to solve problem (\ref{EQ4}),  which is good at handling equality constraints. First,  we introduce an auxiliary variable $\mathbf{G}\in \mathbb{R}^{m \times n}$ and rewrite Eq. (\ref{EQ4}) as 
\begin{equation}
	\begin{gathered}
		\min _{\mathbf{W,O,G}} \frac{1}{2}\left\|\mathbf{W} \mathbf{X}^{\top}-\mathbf{D}\right\|_F^2+\frac{1}{2}\left\|\mathbf{W} \mathbf{X}^{\top} \mathbf{O}-\mathbf{L}\right\|_F^2+\\
		\alpha\left\|\mathbf{G}\right\|_{*}  
		+\lambda\left(\|\mathbf{W}\|_F^2+\|\mathbf{O}\|_F^2\right)\\
		\text{s.t. }  \mathbf{W}\mathbf{X}^{\top} \mathbf{O}=\mathbf{G}.
	\end{gathered}
	\label{EQ5}
\end{equation}
We introduce the augmented Lagrangian function for Eq. (\ref{EQ5}) 
\begin{equation*}
	\begin{gathered}
		\min _{\mathbf{W,O,G}} \frac{1}{2}\left\|\mathbf{W} \mathbf{X}^{\top}-\mathbf{D}\right\|_F^2+\frac{1}{2}\left\|\mathbf{W} \mathbf{X}^{\top} \mathbf{O}-\mathbf{L}\right\|_F^2+
		\alpha\left\|\mathbf{G}\right\|_{*}  \\
		+\lambda\left(\|\mathbf{W}\|_F^2+\|\mathbf{O}\|_F^2\right)+\frac{\mu}{2}\left\|\mathbf{G}-\mathbf{W} \mathbf{X}^{\top} \mathbf{O}-\frac{\mathbf{\Gamma}_1}{\mu}\right\|_F^2,
	\end{gathered}
\end{equation*}
where $\mu$ is a positive penalty parameter, and $\mathbf{\Gamma}_1$ denotes the Lagrangian multipliers. It can be solved by alternately optimizing three sub-problems as follows. The whole process is summarized in Algorithm \ref{A1}. 

\subsubsection{Solving $\mathbf{G}$-Subproblem}
The subproblem w.r.t. $\mathbf{G}$ is
\begin{equation*}
	\mathbf{G}^{k+1}=\underset{\mathbf{G}}{\operatorname{argmin}} \alpha\|\mathbf{G}\|_*+\frac{\mu}{2}\left\|\mathbf{G-W}\mathbf{X}^{\top} \mathbf{O}-\frac{\mathbf{\Gamma}_1}{\mu}\right\|_F^2 .
	\label{G-SUB}
\end{equation*}
It is a  nuclear norm minimization problem and has a closed-form solution \cite{cai2010singular}:
\begin{equation}
	\mathbf{G}^{k+1}=S_{\alpha / \mu}(T),
	\label{GSLOVE}
\end{equation}
where $T=\mathbf{WX}^{\top} \mathbf{O}+\frac{\mathbf{\Gamma}_1}{\mu}$, and ${S(\cdot)}$ is single value thresholding operator. It first performs singular value decomposition on $\mathbf{WX}^{\top} \mathbf{O}+\frac{\mathbf{\Gamma}_1}{\mu}=\mathbf{U} \hat{\mathbf{\Sigma}} \mathbf{V}^{\top}$, and then gives the solution as $\mathbf{U} \hat{\mathbf{\Sigma}} \mathbf{V}^{\top}$, where $\hat{\Sigma}_{i i}=\max \left(0, \Sigma_{i i}-\alpha / \mu\right)$.

\subsubsection{Solving $\mathbf{W}$-Subproblem}
The subproblem w.r.t. $\mathbf{W}$ is
\begin{equation*}
\small
	\begin{aligned}
		\mathbf{W}^{k+1} &=  \underset{\mathbf{W}}{\operatorname{argmin}} \frac{1}{2}\left\|\mathbf{WX}^{\top}-\mathbf{D}\right\|_F^2+\frac{1}{2}\left\|\mathbf{WX}^{\top} \mathbf{O-L}\right\|_T^2 \\
		& +\lambda\|\mathbf{W}\|_F^2+\frac{\mu}{2}\left\|\mathbf{G-WX}^{\top}\mathbf{O}-\frac{\mathbf{\Gamma}_1}{\mu}\right\|_F^2
	\end{aligned}
\end{equation*}
which is  a quadratic optimization problem. The optimal solution is obtained by setting the derivative to zero and equals
\begin{equation}
	\small
	\begin{split}
		\mathbf{W} = & \left(\mathbf{X}^{\top} \mathbf{X}+2 \lambda+\mu \mathbf{X}^{\top} \mathbf{OO}^{\top} \mathbf{X}+\mathbf{X}^{\top} \mathbf{OO}^{\top}\mathbf{X}\right)^{-1}\\
		&(\mu\mathbf{GO}^{\top}\mathbf{X}-\mathbf{\Gamma}_1 \mathbf{O}^{\top} \mathbf{X}+\mathbf{LO}^{\top} \mathbf{X}+\mathbf{DX})
	\end{split}
	\label{WSLOVE}
\end{equation}

\subsubsection{Solving $\mathbf{O}$-Subproblem}
The subproblem w.r.t. $\mathbf{O}$ is
\begin{equation}
	\begin{aligned}
		\mathbf{O}^{k+1} & =  \frac{1}{2}\left\|\mathbf{WX}^{\top} \mathbf{O}-\mathbf{L}\right\|_F^2+\lambda\|\mathbf{O}\|_F^2 \\
		& +\frac{\mu}{2}\left\|\mathbf{G-WX}^{\top} \mathbf{O}-\frac{\mathbf{\Gamma}_1}{\mu}\right\|_F^2
	\end{aligned}
\end{equation}
which is a quadratic optimization problem. The optimal solution is obtained by setting the derivative to zero and equals
\begin{equation}
	\begin{aligned}
		\mathbf{O}^{k+1}= & \left(\mathbf{XW}^{\top} \mathbf{WX}^{\top}+2 \lambda+\mu \mathbf{XW}^{\top}\mathbf{WX}^{\top}\right)^{-1} \\
		+ & \left(\mathbf{XW}^{\top} \mathbf{L}+\mu \mathbf{XW}^{\top}\mathbf{G}-\mathbf{XW}^{\top} \mathbf{\Gamma}_1\right)
	\end{aligned}
	\label{oslove}
\end{equation}

\subsubsection{Updating Multipliers and Penalty Parameter}
Finally, the Lagrange multiplier matrix and penalty parameter are updated based on following rules:
\begin{equation}
	\left\{\begin{array}{l}
		\mathbf{\Gamma}_1^{k+1}=\mathbf{\Gamma}_1^k+\mu^k\left(\mathbf{G}^{k+1}-\mathbf{W}^{k+1} \mathbf{X}^{\top} \mathbf{O}^{k+1}\right) \\
		\mu^{k+1}=\min \left(1.1 \mu, \mu_{\max }\right)
	\end{array}\right.
	\label{chenzislove}
\end{equation}
where $\mu_{max }$ is  the maximum value of $\mu$.

\begin{algorithm}
	\caption{ The proposed methods}
	\label{algorithm: }
	\textbf{Input}: training set $\mathbf{T}$, parameters $\alpha$, $\lambda$, and an instance $\mathbf{x}$ \\
	\textbf{Output}: prediction for $\mathbf{x}$
	\begin{algorithmic}[1]
		\STATE Transforming label distribution into multi-label
		\STATE Initialize $\mathbf{W}$, $\mathbf{O}$, $\mathbf{G}$, $\boldsymbol{\Gamma}_1$, and $\mu$
		\REPEAT
		\STATE Update $\mathbf{G}$ according to Eq. (\ref{GSLOVE})
		\STATE Update $\mathbf{W}$ according to Eq. (\ref{WSLOVE})  
		\STATE Update $\mathbf{O}$ according to Eq. (\ref{oslove})
		\STATE Update $\boldsymbol{\Gamma}_1$ and $\mu$ according to Eq. (\ref{chenzislove})
		\UNTIL{convergence}
		\RETURN $\mathbf{d}^*=\mathbf{W}\mathbf{x}$. 
	\end{algorithmic}
	\label{A1}
\end{algorithm}

\section{Experiments}

\subsection{Experimental Configuration}

\subsubsection{Experimental Datasets}
The experiments are conducted on 16 real-world datasets with label distribution. The key characteristics of these datasets are summarized in Table \ref{tab_dataset}. The first 12 datasets are collected by Geng \cite{geng2016label}. Among these, the first eight ones (from Spoem to Alpha) are from the clustering analysis of genome-wide expression in Yeast Saccharomyces cerevisiae \cite{eisen1998cluster}. The SJAFFE is collected from JAFFE \cite{lyons1998coding}, and the SBU\_3DFE is obtained from BU\_3DFE \cite{yin20063d}. The Gene is obtained from the research on the relationship between genes and diseases \cite{yu2012discriminate}. The Scene consists of multi-label images, where the label distributions are transformed from rankings \cite{geng2014head}. Besides, the SCUT-FBP, M2B, and fbp5500 are about facial beauty perception \cite{ren2017sense}. The last one RAF-ML is a facial expression dataset \cite{li2019blended}. 

\begin{table}
	\renewcommand{\arraystretch}{1.1}
	\centering
	\small
	\begin{tabular}{ccccc}
		\toprule
		ID & Data sets   & \#\textit{Examples} & \#\textit{Features} & \#\textit{Labels}    \\ 
		\midrule
		1     & Spoem   & 2465     & 24       & 2                       \\ 
		2     & Spo5   & 2465     & 24       & 3                       \\ 
		3     & Heat  & 2465     & 24       & 6                       \\ 
		4     & Elu  & 2465     & 24       & 14                       \\ 
		5     & Dtt  & 2465     & 24       & 4                       \\ 
		6     & Cold & 2465     & 24       & 4                       \\ 
		7     & Cdc   & 2465     & 24       & 15                      \\ 
		8     & Alpha & 2465     & 24       & 18                      \\ 
		9     & SJAFFE   & 213    & 243       & 6                      \\ 
		10    & SBU-3DFE     &  2500      & 243      & 6                       \\ 
		11   &  Gene    & 17892     & 36      &  68                       \\ 
		12    & Scene    & 2000     & 294      & 9                       \\
		13    & SCUT-FBP  & 1500     &  300       & 5                       \\ 
		14    & M2B    &  1240     & 250     & 5                       \\ 
		15   & fbp5500   & 5500     & 512      & 5                       \\ 
		16    & RAF-ML    &   4908     & 200      &6                       \\ 
		\bottomrule
	\end{tabular}
	\caption{Details of the dataset.}
	\label{tab_dataset}
\end{table}

\subsubsection{Evaluation Metrics}
We adopt six metrics to evaluate the performance of LDL methods, including Chebyshev ($\downarrow$), Clark ($\downarrow$), Kullback-Leibler (KL) ($\downarrow$), Canberra ($\downarrow$), Intersection ($\uparrow$), and Cosine ($\uparrow$) \cite{geng2016label}. Here, $\downarrow$ indicates that smaller values are better, and $\uparrow$ indicates that larger values are better.

\subsubsection{Comparing Methods}
We compare the proposed methods with seven LDL methods, including IIS-LDL, LDLLDM, EDL-LRL, IncomLDL, Adam-LDL-SCL, LCLR, and LDLLC, which are briefly introduced as follows: 
\begin{itemize}
	\item IIS-LDL \cite{geng2016label}: It  utilizes the maximum entropy model and KL divergence to learn the label distribution and does not consider label correlation. 
 
	\item LDLLDM \cite{wang2021label_2}: It learns the global and local label distribution manifolds to exploit label correlations and can handle incomplete label distribution learning. 
	
	\item  EDL-LRL \cite{jia2019label}: It  captures the low-rank structure locally when learning the label distribution to exploit local label correlations.
	
	\item IncomLDL \cite{xu2017incomplete}: It  utilizes trace-norm regularization and the alternating direction method of multiplier to exploit low-rank label correlation.
	
	\item  Adam-LDL-SCL \cite{jia2021label}: It incorporates local label correlation by encoding it as additional features and simultaneously learns the label distribution and label correlation encoding.
	
	\item  LCLR \cite{ren2019label}: It first models global label correlation using a low-rank matrix and then updates the matrix on clusters of samples to leverage local label correlation.
	
	\item LDLLC \cite{zheng2018label}: LDLLC leverages local label correlation to ensure that prediction distributions between similar instances are as close as possible. 
\end{itemize}

The parameters of the algorithms are as follows.  The suggested parameters are used for IIS-LLD, EDL-LRL, LDLLC, and LDL-LCLR.  For LDLLDM, $\lambda_1, \lambda_2$, and $\lambda_3$ are tuned  from $\left\{10^{-3}, \ldots, 10^3\right\}$, and $g$ is tuned from 1 to 14.  For IncomLDL, $\lambda$ is selected from the range $\left\{2^{-10}, \ldots, 2^{10}\right\}$, and $\rho$ is set to 1. For Adam-LDL-SCL, $\lambda_1, \lambda_2$, and $\lambda_3$ are tuned from the set $\left\{10^{-3}, \ldots, 10^3\right\}$, and $m$ is tuned from 0 to 14. For TLRLDL and TKLRLDL, $\alpha$, $\lambda$ is tuned from $\left\{0.005, 0.01, 0.05, 0.1, 0.5, 1, 10\right\}$, $T$ is selected from 0.1 to 0.5, and $k$ is tuned from 0 to $m$. We run each method for ten-fold cross-validation and tune the parameters on training set. 

\subsection{Results and Discussion}
Table \ref{zhushiyan}  presents the experimental results (mean$\pm$std) of the LDL algorithms on all datasets in terms of Clark, KL, and Cosine (due to limited space, the results in terms of other metrics are reported in the supplementary material), with the best results highlighted in boldface. Moreover, the last row summarized the top-one times of each method.

\begin{figure*}[!h]
	\centering
	\subfloat[KL$\downarrow$]{
		\includegraphics[width=0.3\linewidth]{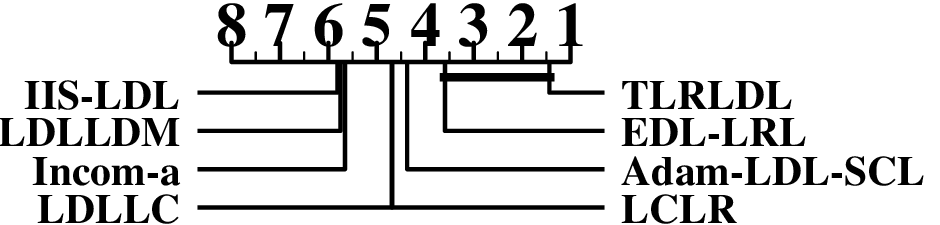}
	}
	\hfil
	\subfloat[Clark $\downarrow$]{
		\includegraphics[width=0.3\linewidth]{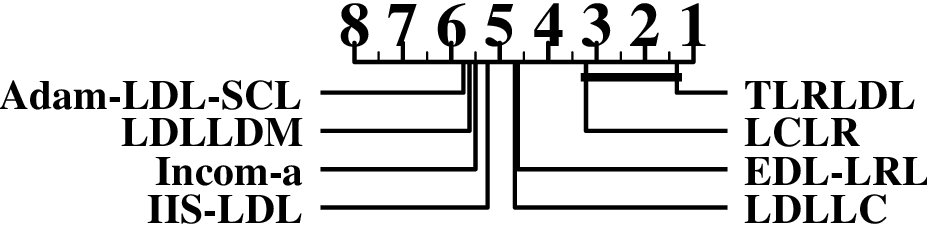}
	}
	\hfil
	\subfloat[Cosine$\uparrow$]{
		\includegraphics[width=0.3\linewidth]{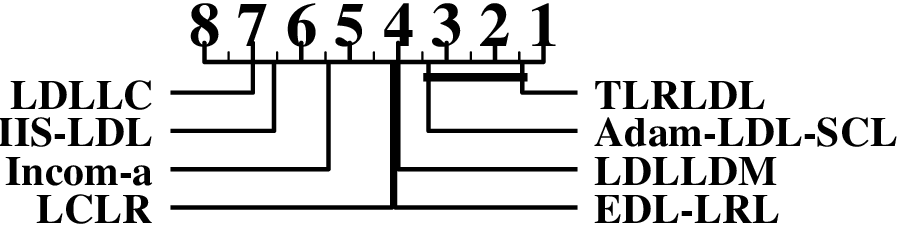}
	}
	\vfil
	\subfloat[Canberra $\downarrow$]{
		\includegraphics[width=0.3\linewidth]{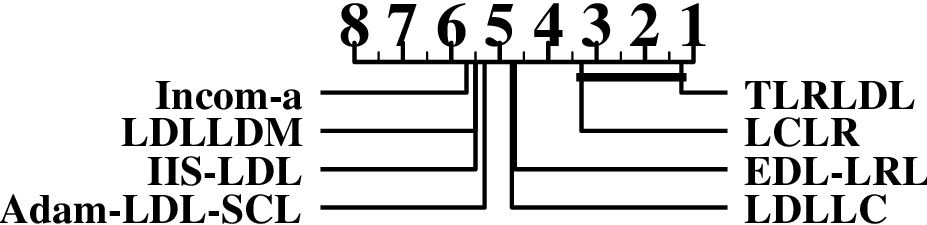}
	}
	\hfil
	\subfloat[Chebyshev$\downarrow$]{
		\includegraphics[width=0.3\linewidth]{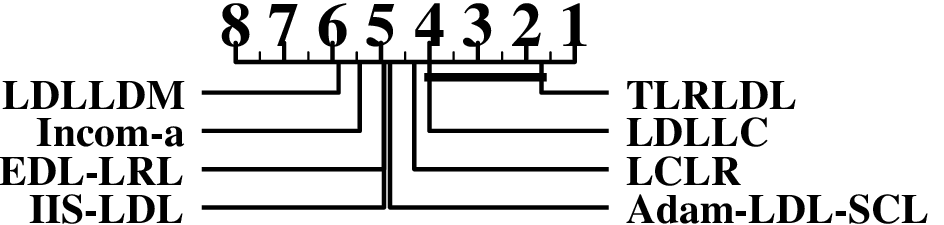}
	}
	\hfil
	\subfloat[Intersection$\uparrow$]{
		\includegraphics[width=0.3\linewidth]{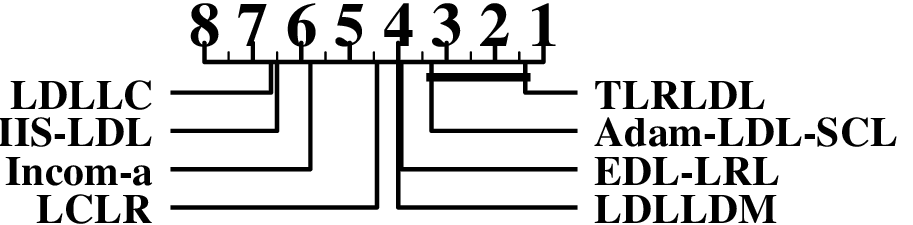}
	}
	
	\caption{CD diagrams of the comparing algorithms in terms of each evaluation criterion. For the tests, CD equals 2.3296 at 0.05 significance level.}
	\label{CD1}
\end{figure*}

 First, we conduct the Friedman test \cite{demvsar2006statistical} to study the comparative performance among the all methods. Table~\ref{criticalFF} shows the Friedman statistics for each metric as well as the critical value. At a confidence level of 0.05, the null hypothesis that \textit{all algorithms achieve equal performance} is clearly rejected. Next, we apply a post-hoc test, i.e., the Bonferroni-Dunn test \cite{demvsar2006statistical},  to compare the relative performance of  TLRLDL against the other algorithms with it as the control algorithm (due to limited pages, the test results with TKLRLDL as the control algorithm are presented in the supplementary material). One algorithm is deemed to achieve significantly different performance from TLRLDL if its average rank differs from that of TLRLDL by at least one critical difference (CD) \cite{demvsar2006statistical}. Figure \ref{CD1} illustrates the CD diagrams for each measure. In each sub-figure, if the average rank of a comparing algorithm is within one CD to that of TLRLDL, they are connected with a thick line; otherwise, it is considered to have a significantly different performance from TLRLDL.

According to Table \ref{zhushiyan}, TLRLDL demonstrates remarkable performance by ranking first in 70.83\% (34 out of 48) of the cases, and it achieves the best mean performance across all metrics. TLRLDL and TKLRLDL achieve the first place in 85.4\% (41 out of 48) of the evaluations, which highlights the effectiveness of our methods. Besides, we can make the following observations from Figure \ref{CD1}:
\begin{itemize}
	\item TLRLDL significantly outperforms IIS-LLD in terms of all metrics. This is because TLRLDL exploits label correlation, but IIS-LLD ignores label correlation, which proves the importance of label correlation for LDL. 
	
	\item TLRLDL achieves significantly better performance than IncomLDL and wins ED-LRL and LCLR by a margin. These three methods all exploit low-rank label correlation in LDL which may not hold as reported in \cite{wang2021label_2}. In comparison, TLRLDL exploits low-rank label correlation in the auxiliary MLL process, which is more appropriate and suitable to LDL. 
	
	\item  Compared with Adam-LDL-SCL, LDLLC, and LDLLDM, TLRLDL also excels, which justifies that the low-rank label correlation on the auxiliary MLL process is a competitive way to consider label correlation for LDL.
\end{itemize}

\begin{table*}
	\centering
	\scriptsize
	\renewcommand{\arraystretch}{1.2}
	\setlength{\tabcolsep}{1.3mm}
	\begin{tabular}{@{}l|l|ccccccccc@{}}
		\toprule
		& Metric & TLRLDL & TKLRLDL & IncomLDL & IIS-LDL & EDL-LRL & Adam-LDL-SCL & LCLR & LDLLC & LDLLDM \\ 
		\midrule
		\multirow{3}{*}{Spoem} & Clark & 0.1238$\pm$.0038 & \textbf{0.1237$\pm$.0197} & 0.1314$\pm$.0028 & 0.1337$\pm$.0014 & 0.1291$\pm$.0000 & 0.1295$\pm$.0000 & 0.1302$\pm$.0001 & 0.1305$\pm$.0014 & 0.1301$\pm$.0303 \\
		& KL & 0.0264$\pm$.0311 & 0.0249$\pm$.0032 & 0.0288$\pm$.1709 & 0.0273$\pm$.0011 & 0.0317$\pm$.0000 & 0.0318$\pm$.0000 & \textbf{0.0246$\pm$.0001} & 0.0254$\pm$.0007 & 0.0264$\pm$.0061 \\
		& Cosine & \textbf{0.9794$\pm$.0007} & 0.9801$\pm$.0028 & 0.9769$\pm$.0239 & 0.9773$\pm$.0005 & 0.9789$\pm$.0000 & 0.9789$\pm$.0000 & 0.9783$\pm$.0003 & 0.9785$\pm$.0005 & 0.9772$\pm$.0071 \\\hline
		\multirow{3}{*}{Spo5} & Clark & \textbf{0.1769$\pm$.0810} & 0.1803$\pm$.0139 & 0.2027$\pm$.0046 & 0.1896$\pm$.0025 & 0.1853$\pm$.0000 & 0.1843$\pm$.0000 & 0.1893$\pm$.0007 & 0.1908$\pm$.0003 & 0.1860$\pm$.0390 \\
		& KL & \textbf{0.0292$\pm$.0343} & 0.0304$\pm$.0408 & 0.0376$\pm$.0148 & 0.0336$\pm$.0008 & 0.0362$\pm$.0000 & 0.0356$\pm$.0000 & 0.0309$\pm$.0000 & 0.0314$\pm$.0000 & 0.0298$\pm$.0336 \\
		& Cosine & \textbf{0.9759$\pm$.0213} & 0.9749$\pm$.0296 & 0.9700$\pm$.0520 & 0.9722$\pm$.0007 & 0.9738$\pm$.0000 & 0.9741$\pm$.0000 & 0.9725$\pm$.0001 & 0.9722$\pm$.0000 & 0.9737$\pm$.0301 \\\hline
		\multirow{3}{*}{Heat} & Clark & \textbf{0.1790$\pm$.0096} & 0.1809$\pm$.0056 & 0.1940$\pm$.0768 & 0.1998$\pm$.0014 & 0.1831$\pm$.0000 & 0.1826$\pm$.0000 & 0.1874$\pm$.0032 & 0.2717$\pm$.0064 & 0.1848$\pm$.0040 \\
		& KL & \textbf{0.0122$\pm$.0279} & 0.0128$\pm$.0005 & 0.0146$\pm$.0141 & 0.0155$\pm$.0002 & 0.0153$\pm$.0000 & 0.0153$\pm$.0000 & 0.0130$\pm$.0003 & 0.0302$\pm$.0020 & 0.0131$\pm$.0002 \\
		& Cosine & \textbf{0.9884$\pm$.0005} & 0.9882$\pm$.0005 & 0.9865$\pm$.0102 & 0.9855$\pm$.0002 & 0.9879$\pm$.0000 & 0.9880$\pm$.0000 & 0.9878$\pm$.0002 & 0.9695$\pm$.0022 & 0.9875$\pm$.0003 \\\hline
		\multirow{3}{*}{Elu} & Clark & 0.2028$\pm$.0024 & \textbf{0.2000$\pm$.0103} & 0.2325$\pm$.0612 & 0.2395$\pm$.0022 & 0.1998$\pm$.0000 & 0.1989$\pm$.0000 & 0.2032$\pm$.0028 & 0.4114$\pm$.0089 & 0.2010$\pm$.0011 \\
		& KL & \textbf{0.0062$\pm$.0706} & 0.0063$\pm$.0138 & 0.0066$\pm$.0123 & 0.0091$\pm$.0002 & 0.0073$\pm$.0000 & 0.0072$\pm$.0000 & 0.0064$\pm$.0001 & 0.0296$\pm$.0011 & 0.0063$\pm$.0003 \\
		& Cosine & \textbf{0.9941$\pm$.0010} & 0.9940$\pm$.0077 & 0.9918$\pm$.0049 & 0.9911$\pm$.0002 & 0.9940$\pm$.0000 & 0.9940$\pm$.0000 & 0.9938$\pm$.0001 & 0.9667$\pm$.0013 & 0.9940$\pm$.0005 \\\hline
		\multirow{3}{*}{Cdc} & Clark & 0.2142$\pm$.0719 & \textbf{0.2094$\pm$.0127} & 0.2243$\pm$.0016 & 0.2537$\pm$.0026 & 0.2168$\pm$.0000 & 0.2161$\pm$.0000 & 0.2172$\pm$.0021 & 0.4259$\pm$.0013 & 0.2147$\pm$.0309 \\
		& KL & 0.0073$\pm$.0061 & 0.0070$\pm$.0017 & 0.0080$\pm$.0765 & 0.0099$\pm$.0002 & 0.0082$\pm$.0000 & 0.0082$\pm$.0000 & 0.0072$\pm$.0002 & 0.0291$\pm$.0001 & \textbf{0.0068$\pm$.0338} \\
		& Cosine & \textbf{0.9935$\pm$.0263} & 0.9934$\pm$.0016 & 0.9926$\pm$.0300 & 0.9905$\pm$.0002 & 0.9933$\pm$.0000 & 0.9933$\pm$.0000 & 0.9932$\pm$.0002 & 0.9680$\pm$.0003 & 0.9934$\pm$.0463 \\\hline
		\multirow{3}{*}{Dtt} & Clark & \textbf{0.0946$\pm$.0175} & 0.0975$\pm$.0013 & 0.1039$\pm$.0188 & 0.1162$\pm$.0009 & 0.0993$\pm$.0000 & 0.0986$\pm$.0000 & 0.0971$\pm$.0006 & 0.1738$\pm$.0011 & 0.0959$\pm$.0003 \\
		& KL & 0.0060$\pm$.0076 & 0.0062$\pm$.0008 & 0.0066$\pm$.0765 & 0.0088$\pm$.0002 & 0.0098$\pm$.0000 & 0.0098$\pm$.0000 & \textbf{0.0059$\pm$.0000} & 0.0223$\pm$.0005 & 0.0059$\pm$.0012 \\
		& Cosine & \textbf{0.9945$\pm$.0132} & 0.9942$\pm$.0013 & 0.9933$\pm$.0300 & 0.9916$\pm$.0001 & 0.9940$\pm$.0000 & 0.9940$\pm$.0000 & 0.9943$\pm$.0000 & 0.9783$\pm$.0004 & 0.9944$\pm$.0584 \\\hline
		\multirow{3}{*}{Alpha} & Clark & \textbf{0.2072$\pm$.0042} & 0.2079$\pm$.0314 & 0.2156$\pm$.0775 & 0.2585$\pm$.0015 & 0.2107$\pm$.0000 & 0.2103$\pm$.0000 & 0.2085$\pm$.0012 & 0.4501$\pm$.0019 & 0.2116$\pm$.0236 \\
		& KL & \textbf{0.0052$\pm$.0016} & 0.0054$\pm$.0060 & 0.0058$\pm$.0232 & 0.0084$\pm$.0001 & 0.0063$\pm$.0000 & 0.0063$\pm$.0000 & 0.0054$\pm$.0001 & 0.0267$\pm$.0002 & 0.0055$\pm$.0857 \\
		& Cosine & \textbf{0.9948$\pm$.0015} & 0.9947$\pm$.0069 & 0.9943$\pm$.0052 & 0.9916$\pm$.0001 & 0.9946$\pm$.0000 & 0.9946$\pm$.0000 & 0.9947$\pm$.0001 & 0.9700$\pm$.0001 & 0.9946$\pm$.0362 \\\hline
		\multirow{3}{*}{Cold} & Clark & 0.1378$\pm$.0014 & 0.1390$\pm$.0731 & 0.1463$\pm$.0338 & 0.1568$\pm$.0014 & 0.1403$\pm$.0000 & 0.1398$\pm$.0000 & 0.1416$\pm$.0042 & 0.1512$\pm$.0040 & \textbf{0.1363$\pm$.0190} \\
		& KL & 0.0118$\pm$.0042 & 0.0124$\pm$.0565 & 0.0139$\pm$.0056 & 0.0153$\pm$.0002 & 0.0162$\pm$.0000 & 0.0162$\pm$.0000 & 0.0128$\pm$.0009 & 0.0140$\pm$.0006 & \textbf{0.0116$\pm$.0154} \\
		& Cosine & \textbf{0.9892$\pm$.0061} & 0.9886$\pm$.0357 & 0.9873$\pm$.0047 & 0.9855$\pm$.0003 & 0.9885$\pm$.0000 & 0.9885$\pm$.0000 & 0.9880$\pm$.0008 & 0.9866$\pm$.0005 & 0.9889$\pm$.0390 \\\hline
		\multirow{3}{*}{SJA} & Clark & \textbf{0.3602$\pm$.0042} & 0.3657$\pm$.0099 & 0.4567$\pm$.0061 & 0.4516$\pm$.0181 & 0.4232$\pm$.0002 & 1.3730$\pm$.9671 & 0.4049$\pm$.0082 & 0.4369$\pm$.0034 & 0.4153$\pm$.0010 \\
		& KL & \textbf{0.0480$\pm$.0016} & 0.0518$\pm$.0277 & 0.0659$\pm$.0202 & 0.0790$\pm$.0053 & 0.0692$\pm$.0000 & 1.0106$\pm$.9024 & 0.0663$\pm$.0000 & 0.0791$\pm$.0019 & 0.0668$\pm$.0009 \\
		& Cosine & \textbf{0.9558$\pm$.0015} & 0.9509$\pm$.0528 & 0.9321$\pm$.0187 & 0.9208$\pm$.0062 & 0.9319$\pm$.0000 & 0.6503$\pm$.0850 & 0.9372$\pm$.0000 & 0.9245$\pm$.0019 & 0.9363$\pm$.0125 \\\hline
		\multirow{3}{*}{SCU} & Clark & \textbf{1.0793$\pm$.0061} & 1.4568$\pm$.0000 & 1.5459$\pm$.0016 & 1.5007$\pm$.0064 & 1.5146$\pm$.0000 & 1.4654$\pm$.0000 & 1.3859$\pm$.0062 & 2.6438$\pm$.0000 & 1.3978$\pm$.0009 \\
		& KL & 0.1779$\pm$.0015 & \textbf{0.1503$\pm$.0528} & 2.6539$\pm$.0221 & 0.1824$\pm$.0170 & 9.2314$\pm$.0504 & 7.4655$\pm$.0041 & 0.4248$\pm$.0047 & 16.040$\pm$.1750 & 0.3997$\pm$.0009 \\
		& Cosine & 0.8208$\pm$.0028 & 0.7436$\pm$.0202 & 0.6108$\pm$.0689 & 0.6627$\pm$.0028 & 0.6477$\pm$.0000 & 0.7436$\pm$.0000 & 0.8126$\pm$.0011 & 0.5144$\pm$.0002 & \textbf{0.8375$\pm$.0002} \\\hline
		\multirow{3}{*}{SBU} & Clark & \textbf{0.3455$\pm$.0043} & 0.3520$\pm$.0016 & 0.3692$\pm$.0011 & 0.4217$\pm$.0029 & 0.4061$\pm$.0000 & 0.3718$\pm$.0000 & 0.3956$\pm$.0039 & 0.4172$\pm$.0003 & 0.4056$\pm$.0071 \\
		& KL & \textbf{0.0502$\pm$.0857} & 0.0552$\pm$.0765 & 0.0619$\pm$.0036 & 0.0776$\pm$.0009 & 0.0726$\pm$.0000 & 0.0604$\pm$.0000 & 0.2008$\pm$.0026 & 0.0845$\pm$.0002 & 0.0791$\pm$.0370 \\
		& Cosine & \textbf{0.9474$\pm$.0012} & 0.9449$\pm$.0300 & 0.9410$\pm$.0013 & 0.9177$\pm$.0011 & 0.9232$\pm$.0000 & 0.9367$\pm$.0000 & 0.9267$\pm$.0011 & 0.9180$\pm$.0002 & 0.9226$\pm$.0010 \\\hline
		\multirow{3}{*}{RAF} & Clark & \textbf{0.8652$\pm$.0082} & 1.4327$\pm$.0528 & 1.5597$\pm$.0234 & 1.5581$\pm$.0086 & 1.4495$\pm$.0003 & 1.4585$\pm$.0000 & 1.5962$\pm$.0138 & 1.6210$\pm$.0034 & 1.4151$\pm$.0016 \\
		& KL & \textbf{0.0864$\pm$1.8674} & 0.2086$\pm$.0187 & 6.4358$\pm$.0090 & 3.5105$\pm$.0654 & 2.2182$\pm$.0013 & 5.6995$\pm$.0000 & 13.7926$\pm$1.8887 & 0.7347$\pm$.0001 & 0.2699$\pm$.0109 \\
		& Cosine & \textbf{0.9252$\pm$.0034} & 0.9234$\pm$.0115 & 0.5631$\pm$.0101 & 0.7351$\pm$.0020 & 0.9198$\pm$.0000 & 0.8706$\pm$.0000 & 0.7968$\pm$.0047 & 0.6453$\pm$.0007 & 0.8976$\pm$.0002 \\\hline
		\multirow{3}{*}{M2B} & Clark & \textbf{1.0224$\pm$.0009} & 1.5160$\pm$.0023 & 1.4832$\pm$.0878 & 1.2282$\pm$.0070 & 1.6770$\pm$.0046 & 1.2093$\pm$.0000 & 1.6902$\pm$.1999 & 1.6791$\pm$.0002 & 1.5538$\pm$.0029 \\
		& KL & \textbf{0.6972$\pm$.0108} & 0.7786$\pm$.0109 & 0.8180$\pm$.0301 & 0.8572$\pm$.0354 & 0.8632$\pm$.0044 & 0.8128$\pm$.0000 & 0.9528$\pm$.5826 & 0.9051$\pm$.0000 & 0.7556$\pm$.0083 \\
		& Cosine & 0.7431$\pm$.0004 & \textbf{0.7786$\pm$.0070} & 0.7583$\pm$.0308 & 0.7588$\pm$.0122 & 0.7423$\pm$.0006 & 0.7639$\pm$.0000 & 0.5786$\pm$.0325 & 0.6039$\pm$.0000 & 0.6953$\pm$.0054 \\\hline
		\multirow{3}{*}{Gene} & Clark & \textbf{2.0077$\pm$.0041} & 2.1086$\pm$.0036 & 2.1110$\pm$.0040 & 2.1734$\pm$.0269 & 2.1102$\pm$.0000 & 2.1144$\pm$.0000 & 2.0677$\pm$.0172 & 2.1162$\pm$.0009 & 2.1374$\pm$.0036 \\
		& KL & \textbf{0.2224$\pm$.0096} & 0.2236$\pm$.0091 & 0.2372$\pm$.0002 & 0.2380$\pm$.0069 & 0.2258$\pm$.0000 & 0.2256$\pm$.0000 & 0.3618$\pm$.0026 & 0.2374$\pm$.0054 & 0.2455$\pm$.0069 \\
		& Cosine & \textbf{0.8387$\pm$.0028} & 0.8376$\pm$.0038 & 0.8342$\pm$.0003 & 0.8274$\pm$.0036 & 0.8347$\pm$.0000 & 0.8345$\pm$.0000 & 0.8374$\pm$.0018 & 0.8338$\pm$.0027 & 0.8290$\pm$.0021 \\\hline
		\multirow{3}{*}{fbp} & Clark & \textbf{0.5510$\pm$.0036} & 0.6288$\pm$.0155 & 1.2938$\pm$.0011 & 1.5065$\pm$.0023 & 1.6994$\pm$.0001 & 1.2755$\pm$.0000 & 1.4102$\pm$.1809 & 1.8756$\pm$.0857 & 1.2747$\pm$.0122 \\
		& KL & 0.0904$\pm$.0091 & \textbf{0.0800$\pm$.0068} & 0.2017$\pm$.0003 & 4.2207$\pm$.0332 & 1.3803$\pm$.0029 & 0.7725$\pm$.0001 & 0.4208$\pm$.4110 & 1.7692$\pm$.4081 & 0.1149$\pm$.0047 \\
		& Cosine & \textbf{0.9567$\pm$.0038} & 0.9500$\pm$.0047 & 0.9412$\pm$.0005 & 0.6572$\pm$.0017 & 0.7943$\pm$.0000 & 0.9521$\pm$.0000 & 0.8527$\pm$.0051 & 0.8242$\pm$.0123 & 0.9528$\pm$.0056 \\\hline
		\multirow{3}{*}{Sce} & Clark & \textbf{2.1786$\pm$.0108} & 2.4959$\pm$.0061 & 2.4765$\pm$.0303 & 2.4685$\pm$.0135 & 2.4348$\pm$.0000 & 2.4665$\pm$.0000 & 2.4230$\pm$.0025 & 2.4784$\pm$.0007 & 2.6316$\pm$.0000 \\
		& KL & 2.1128$\pm$.0061 & 2.2857$\pm$.0015 & \textbf{0.2378$\pm$.0061} & 3.0463$\pm$.0490 & 2.3857$\pm$.0002 & 2.7001$\pm$.0000 & 0.8287$\pm$.0227 & 1.0243$\pm$.0027 & 4.4471$\pm$.0000 \\
		& Cosine & 0.7989$\pm$.0016 & \textbf{0.8346$\pm$.0008} & 0.7297$\pm$.0071 & 0.6614$\pm$.0044 & 0.7273$\pm$.0000 & 0.7163$\pm$.0000 & 0.7290$\pm$.0042 & 0.6446$\pm$.0001 & 0.3603$\pm$.0000 \\ 
		\midrule
		\multicolumn{2}{c|}{top-1 times} & \textbf{34} & 7 & 1 & 0 & 0 & 0 & 2 & 0 & 4 \\
		\bottomrule
	\end{tabular}
	\caption{Results (mean$\pm$std) of the comparing methods in terms of three metrics (more results in terms of other three metrics are reported in the supplementary material) on 16 LDL datasets, where the best results are highlighted in boldface. }
	\label{zhushiyan}
\end{table*}

\begin{table*}[!h]
	\small
	\centering
	\renewcommand{\arraystretch}{1.2}
	\setlength{\tabcolsep}{3.2mm}
	\begin{tabular}{@{}l|ccccccc@{}}
		\hline
		Critical Value ($\alpha=0.05$) & Evaluation metric & Chebyshev & Clark & Canberra & KL & Cosine & Intersection \\ \hline
		\multicolumn{1}{c|}{2.8500} & Friedman Statistics $F_F$ & 28.0098 & 44.9412 & 46.3235 & 38.0000 & 45.7059 & 44.0158 \\ \hline
	\end{tabular}
	\caption{Summary of the Friedman statistics $F_F$ 
		in terms  of six evaluation metrics, as well as the critical value at a significance level of 0.05. }
	\label{criticalFF}
\end{table*}

To summarize, the experimental results validate the competitive performance of the proposed algorithms.

\subsection{Ablation Study}
Next, we study the advantages of exploiting the low-rank label correlation on the auxiliary MLL. First, we derive TLRLDL-a by
\begin{equation*}
		\min _{\mathbf{W}} \frac{1}{2}\left\|\mathbf{W} \mathbf{X}^{\top}-\mathbf{D}\right\|_F^2 + \alpha\left\|\mathbf{W}\mathbf{X}^{\top}\right\|_{*}  
		+\lambda \|\mathbf{W}\|_F^2. 
\end{equation*}
Second, we derive TLRLDL-b by keeping the first and fourth items of Eq. (\ref{EQ4}).  That is, TLRLDL-a exploits low-rank label correlation on LDL, and TLRLDL-b ignores label correlation. We then compare TLRLDL with TLRLDL-a and TLRLDL-b.

\begin{figure*}[!h]
	\centering
	\subfloat[Clark]{
		\includegraphics[scale=0.25]{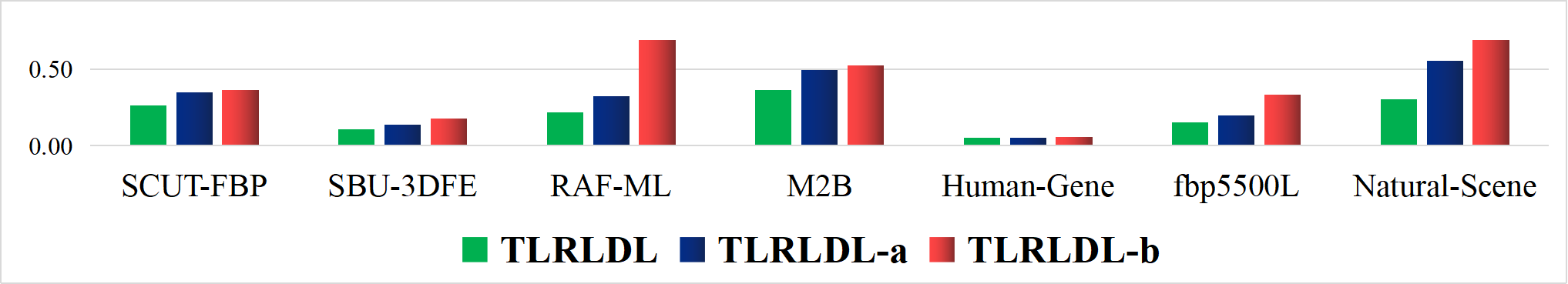}
	}
	\hfil
	\subfloat[KL]{
		\includegraphics[scale=0.25]{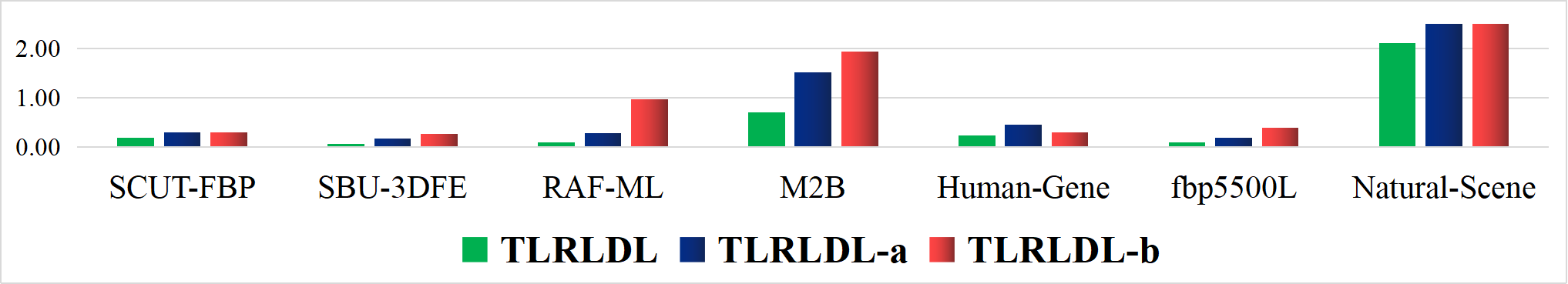}
	}
	\hfil
	\subfloat[Cosine]{
		\includegraphics[scale=0.25]{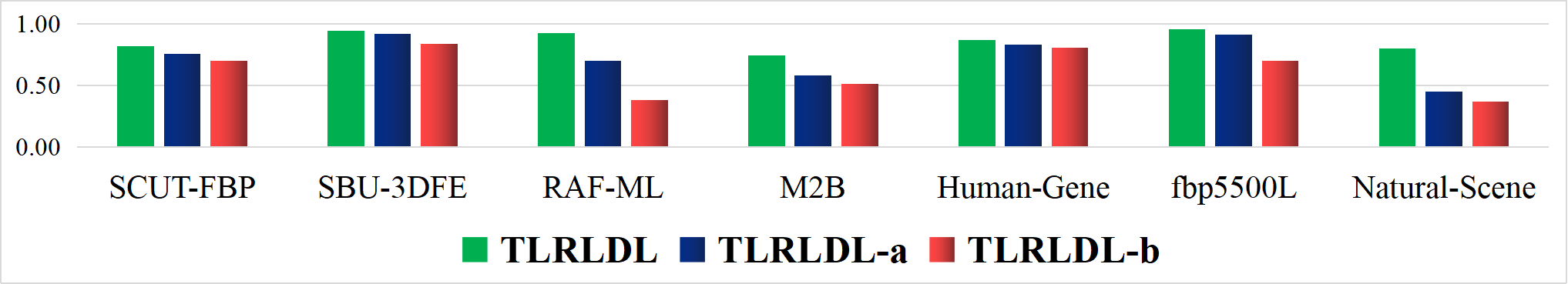}
	}
	\hfil
	\subfloat[Intersection]{
		\includegraphics[scale=0.25]{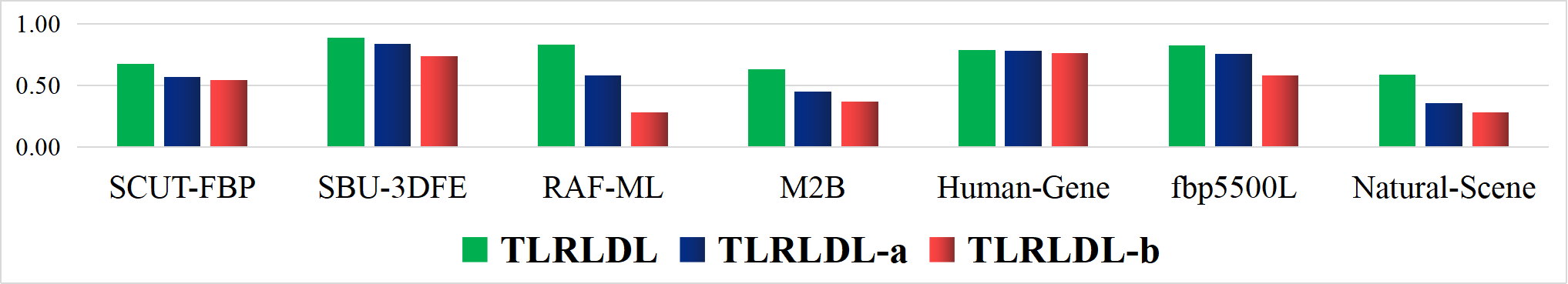}
	}
	\caption{ Ablation results on seven datasets  in terms of Clark $\downarrow$, KL $\downarrow$, Cosine $\uparrow$, and Intersection $\uparrow$.}
	\label{xiaorongkeshihua}
\end{figure*}

Figure \ref{xiaorongkeshihua} presents the comparison results in terms of  Clark, KL, Cosine, and Intersection. Further, we conduct the Wilcoxon signed-rank tests \cite{demvsar2006statistical} for TLRLDL against TLRLDL-a and TLRLDL-b and report the results of the tests in Table \ref{w-test}. According to Figure \ref{xiaorongkeshihua} and Table \ref{w-test}, we can draw the following conclusions: \begin{itemize}
	\item  TLRLDL and TLRLDL-a have better performance than TLRLDL-b. TLRLDL-b ignores label correlation, while TLRLDL and TLRLDL-a consider label correlation, which improves their performance. This observation further justifies the importance of label correlation for LDL. 
	
	\item  TLRLDL significantly outperforms TLRLDL-a. Since the different between TLRLDL and TLRLDL-a lies in that the former (the latter, respectively) exploits low-rank label correlation on MLL (LDL, respectively), this observation clearly justifies benefits of exploits low-rank label correlation on the auxiliary MLL. 
\end{itemize}  

\begin{table*}[!h]
	\renewcommand{\arraystretch}{1.2}
	\small
	\centering
	\setlength{\tabcolsep}{3.5mm}
	\begin{tabular}{@{}ccccccc@{}}
		\hline
		TLRLDL \textit{\textbf{vs}.}& Chebyshev$\downarrow$ & Clark$\downarrow$  & Canberra$\downarrow$  & KL$\downarrow$  & Cosine$\uparrow$  & Intersection $\uparrow$ \\ \hline
		TLRLDL-a & $\mathbf {win}$[4.37e-04] & $\mathbf {win}$[4.38e-04] & $\mathbf {win}$ [4.37e-04]& $\mathbf {win}$[4.46e-03] & $\mathbf {win}$[4.38e-04] & $\mathbf {win}$[4.38e-04] \\
		TLRLDL-b & $\mathbf {win}$[4.38e-04] & $\mathbf {win}$ [3.20e-03]& $\mathbf {win}$[1.61e-03] & $\mathbf {win}$[4.38e-04] & $\mathbf {win}$ [4.38e-04]& $\mathbf {win}$[4.38e-04] \\ \hline
	\end{tabular}
	\caption{The results (Win/Tie/Lose[$p$-value]) of the Wilcoxon signed-rank tests for TLRLDL against TLRLDL-a and TLRLDL-b at a confidence level of 0.05.}
	\label{w-test}
\end{table*}

\subsection{Parameter Sensitivity Analysis}
TLRLDL has two trade-off parameters, including $\alpha$ and $\lambda$. Next, we analyze the sensitivity of them.  

\begin{figure*}[!h]
	\centering
	\centering
	\subfloat[Chebyshev  with varying $\alpha$]{
		\includegraphics[scale=0.28, trim={0cm 0 1.5cm 0}, clip]{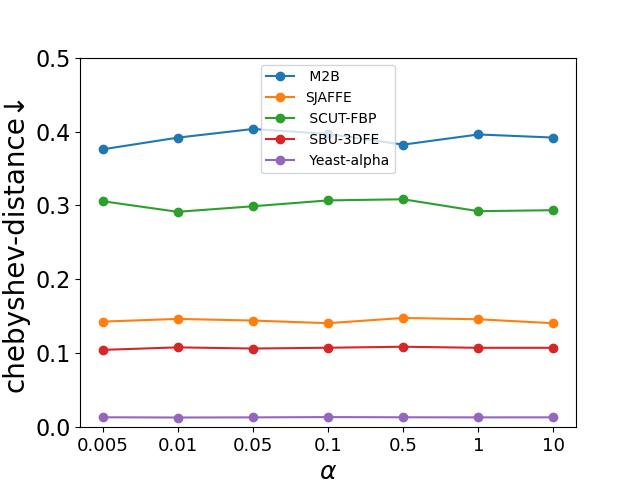}
	}
	\hfil
	\subfloat[Clark with varying $\alpha$]{
		\includegraphics[scale=0.28, trim={0cm 0 1.5cm 0}, clip]{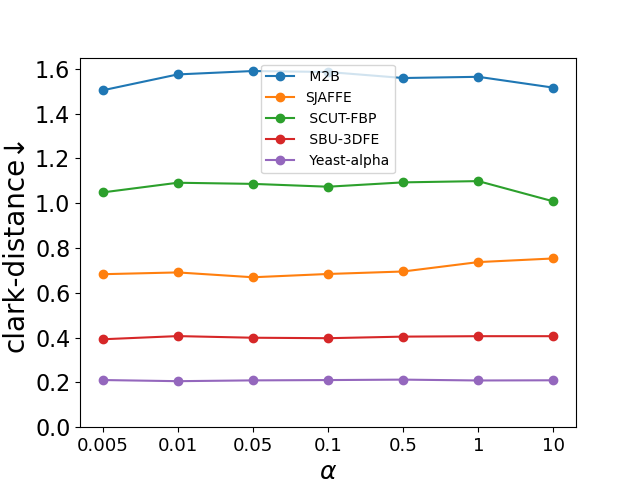}
	}
	\hfil
	\subfloat[Chebyshev with varying $\lambda$]{
		\includegraphics[scale=0.28, trim={0cm 0 1.5cm 0}, clip]{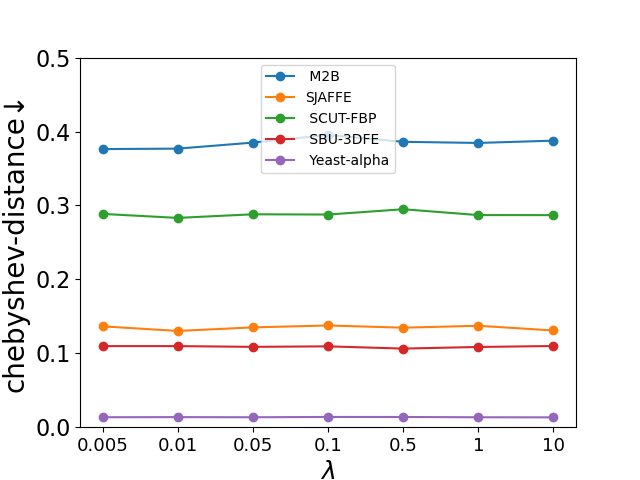}
	}
	\hfil
	\subfloat[Clark with varying $\lambda$]{
		\includegraphics[scale=0.28, trim={0cm 0 1.5cm 0}, clip]{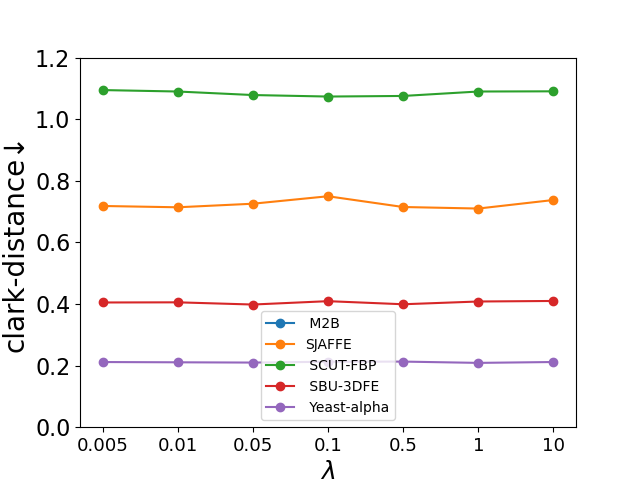}
	}
	\caption{The performance of TLRLDL with $\alpha$ and $\lambda$ varying from $\left\{0.005, 0.01, 0.05, 0.1, 0.5, 1, 10\right\}$ in terms of Chebyshev and Clark on SCUT-FBP, M2B, SJAFFE, SBU\_3DFE, and Alpha.}
	\label{canshufenxi}
\end{figure*}

First, we run TLRLDL with $\alpha$ varying from the candidate set  $\{0.005, 0.01, 0.05, 0.1, 0.5, 1, 10\}$ and report its performance on SCUT-FBP, M2B, SJAFFE, SBU\_3DFE, and Alpha in Figure \ref{canshufenxi}. As can be seen from Figure \ref{canshufenxi}, TLRLDL shows robustness w.r.t. $\alpha$. As a result, we can set $\alpha$ to 0.1 to get satisfying performance. Likewise, we also run TLRLDL with $\lambda$ ranging from the same candidate set and present its performance in Figure \ref{canshufenxi}. From Figure \ref{canshufenxi}, TLRLDL is robust w.r.t. $\lambda$. We may expect satisfying performance for $\lambda=0.1$. 

\section{Conclusion}
LDL has an exponential-sized output space--with a size of $\mathbb{R}^m$--which may decrease the performance of existing algorithms. To solve that, many LDL studies have proposed to exploit label correlation. Among these, some have proposed to exploit low-rank label correlation of label distribution, which may not hold as disclosed by \cite{wang2021label_2} because LDL matrices are typically full-rank. To address this problem, we introduce an auxiliary MLL process in LDL and exploit low-rank label correlation on the MLL process instead of LDL. By doing this, the low-rank label correlation is implicitly exploited in our LDL methods. We conduct extensive experiments and show that our proposed methods achieve remarkably better performance than several state-of-the-art LDL methods. Moreover, further ablation studies justify the advantages of exploiting low-rank label correlation on the auxiliary MLL process. 

We believe our work brings a new perspective for modeling the label correlation for LDL. In future work, we will extend our work to exploit local low-rank label correlation.

\bibliography{aaai24}

\end{document}